\documentclass[runningheads]{llncs}
\usepackage{graphicx}
\usepackage{amsmath,amssymb}
\usepackage{booktabs}

\begin{document}

\title{PIRC Net : Using Proposal Indexing, Relationships and Context for Phrase Grounding} 
\titlerunning{PIRC Net} 


\author{Rama Kovvuri \and
Ram Nevatia}

\authorrunning{Rama Kovvuri, Ram Nevatia.} 


\institute{University of Southern California, Los Angeles, CA 90007, USA\\
\email{\{nkovvuri,nevatia\}@usc.edu}}

\maketitle

\begin{abstract}
Phrase Grounding aims to detect and localize objects in images that are referred to and are queried by natural language phrases. Phrase grounding finds applications in tasks such as Visual Dialog, Visual Search and Image-text co-reference resolution. In this paper, we present a framework that leverages information such as phrase
category, relationships among neighboring phrases in a sentence and context to improve the performance of phrase grounding systems. We propose three modules: Proposal Indexing Network(PIN); Inter-phrase Regression Network(IRN) and Proposal Ranking Network(PRN) each of which analyze the region proposals of an image at increasing levels of detail by incorporating the above information. Also, in the absence of ground-truth spatial locations of the phrases(weakly-supervised), we propose knowledge transfer mechanisms that leverages the framework of PIN module. We demonstrate the effectiveness of our approach on the Flickr 30k Entities and ReferItGame datasets, for which we achieve improvements over state-of-the-art approaches in both supervised and weakly-supervised variants.

\end{abstract}
\section{Introduction}

Our goal is to detect and localize objects in images that are referred to, are queried by, natural language phrases. This objective is also commonly referred to as ``phrase grounding". The key difference with conventional object detection approaches is that the categories of objects is not pre-specified; furthermore the queries may contain attributes (such as a``red car")  and relations between objects (``baby holding a pacifier"). Phrase grounding finds applications in tasks such as Visual Dialog~\cite{cvpr2017dialog}, Visual Search~\cite{gordo2016deep} and Image-text co-reference resolution~\cite{rohrbach2017cvpr}.

 Grounding faces several challenges beyond those present for learning detectors for pre-specified categories. These include generalizing the models from limited data, resolving semantic ambiguities from an open-ended vocabulary and localizing small, hard-to-detect visual entities. Current approaches (~\cite{rohrbach2016grounding}~\cite{hu2016natural}~\cite{plummer2015flickr30k}~\cite{myiccv2017}) adopt a two stage process for phrase grounding. In the first stage, a region proposal module~\cite{uijlings2013selective}~\cite{zitnick2014edge} generates proposals to identify regions likely to contain objects or groups of objects in an image. In the second stage, the grounding system employs a multimodal subspace~\cite{ourWACV} that projects queries (textual modality) and their corresponding proposals (visual modality) to have high correlation score. 
 
  Various approaches are suggested to learn this subspace such as knowledge transfer from image captioning~\cite{hu2016natural}; Canonical Correlation Analysis~\cite{plummer2015flickr30k}~\cite{wang2016deepcca} and query-specific attention for region proposals~\cite{rohrbach2016grounding}~\cite{myiccv2017}. To provide visual context, ~\cite{nagaraja2016modeling} augments the proposals' visual features using bounding box features while ~\cite{yu2016modeling} employs neighboring proposal features. Query phrases are often generated from image descriptions and neighboring phrases from a description can provide useful context like relative locations and relationships between them to reduce semantic ambiguity. Using this context, ~\cite{plummer2017sppc}~\cite{kan2017msrc} ground multiple queries for the same image together to resolve ambiguity and reduce conflicting predictions.
\begin{figure}[h]
\vspace*{-0.75cm}
\includegraphics[height=1.95in, width=0.95\textwidth]{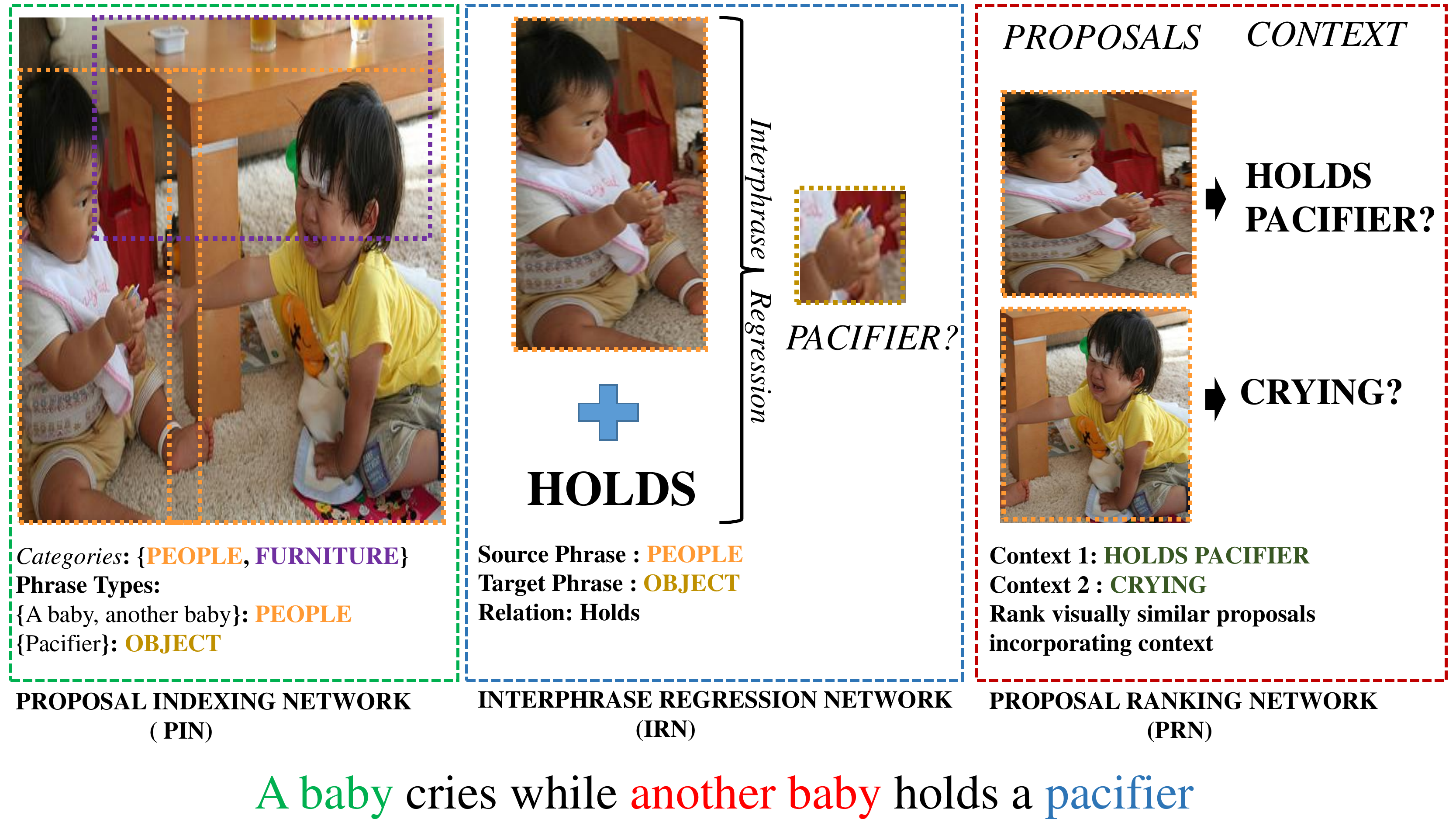}
\vspace*{-0.35cm}
\centering
\caption{Overview of the PIRC Net framework}\label{fig: ovrvw}
\vspace*{-0.85cm}
\end{figure}

Given a query phrase, above approaches consider all the proposals extracted from the image for grounding. This can result in inter-class errors especially for object classes with few corresponding query phrases in the training dataset. The existing approaches are also upperbound by the accuracy of proposal generator, since a query phrase is only able to choose from pre-generated proposals. Further, the existing approaches provide bounding box location and/or image visual representation as context for a proposal. However, many of the queries are defined by attributes and the suitability of an attribute to a proposal is relative. For example, for a query phrase ``short boy" we need to compare each boy with all other boys in the image to figure out which one is the ``short boy".

To address the above challenges, we present a framework that uses Proposal Indexing, Relationship and Context for Grounding (``PIRC Net"), whose overview is presented in Figure \ref{fig: ovrvw}. We propose an architecture with three modules; each succeeding module analyzes the region proposals for a query phrase at increasing levels of detail. The first module, Proposal Indexing Network (PIN) is a two stage network that is trained to reduce the generalization errors and volume of proposals required per query. The second module, Inter-phrase Regression Network (IRN) functions as an enhancement module by generating positive proposals for the cases where PIN does not generate a positive proposal. The third and final module, Proposal Ranking Network (PRN) scores the proposals generated by PIN and IRN by incorporating query specific context. A brief description of each of the modules is provided below. 

In the first stage, PIN module classifies the region proposals into higher-level phrase categories. The phrase categories are generated by grouping semantically similar query phrases together using clustering. Intuitively, if the training data has no query phrases from object class ``bus", queries such as ``red bus" should consider the proposals that have similar appearance to other queries from same phrase category like ``sports car" and ``yellow truck". Skip-thought vectors~\cite{Kiros2015Skip} are used to encode the phrases since they latently identify the important nouns in a phrase to map semantically similar phrases. In the next stage, PIN learns to attend the proposals most relevant to query phrase to further reduce the volume of region proposals. IRN module employs a novel method estimate the location of one neighbor phrase from another given the relationship between them. For example, for a phrase tuple {'small baby', 'holds', 'a pacifier'} while it is difficult to detect 'a pacifier' alone, the relationship 'holds' can help localize 'a pacifier' from the location of its neighboring phrase 'small baby'. PRN module trains a bimodal network which uses learned embeddings to score the region proposals (visual modality) given a query phrase (textual modality). To encode the visual context given a query phrase, we compare a region proposal to other proposals from the same phrase category by encoding their relative appearance.

During training, IRN and PRN learn to predict the proposals with high overlap to ground truth bounding box. However, the ground truth annotation is costly and most vision and language datasets do not provide this information. While it is difficult to train IRN and PRN without this information, we provide methodologies to train PIN with ground truth (supervised) and without ground truth (weakly-supervised) annotations. For supervised setting, PIN is trained using a RPN to predict the proposals close to the groundtruth bounding boxes for a phrase category. For the weakly supervised setting, we propose to use knowledge transfer learning from an object detection system for training PIN and retrieving region proposals that may belong to query phrase category.

We evaluate our framework on two common phrase grounding datasets: Flickr 30K entities~\cite{plummer2015flickr30k} and Refer-it Game~\cite{KazemzadehOrdonezMattenBergEMNLP14}. For supervised setting, experiments show that our framework outperforms existing state-of-the-art~\cite{myiccv2017}, achieving 6\%/8\% and 10\%/15\% improvements using VGG/ResNet architectures on Flickr30K and Referit datasets respectively. For weakly-supervised setting, our framework achieves 5\% and 4\% improvements over state-of-the-art~\cite{rohrbach2016grounding} for both datasets.

Our contributions are: (a) Designed a query-guided Proposal Indexing Network that reduces generalization errors for grounding; (b) Introduced novel Inter-phrase Regression and Proposal Ranking Networks that leverage the context provided by multiple phrases in a caption; (c) Proposed knowledge transfer mechanisms that employ object detection systems to index proposals in weakly supervised setting.


\section{Related Work}

\textbf{Phrase grounding} Improving hugely on early phrase grouding attempts that used limited vocabulary~\cite{karpathy2014deep}~\cite{kong2014cvpr}, Karpathy et al~\cite{karpathy2015deep} employ bidirectional RNN to align the sentence fragments and image regions in common embedding space. Hu et al~\cite{hu2016natural} proposed to rank proposals using knowledge transfer from image captioning. Rohrbach et al~\cite{rohrbach2016grounding} employ attention to rank proposals in a latent subspace. Chen et al~\cite{kan2017msrc} extended this approach to account for regression based on query semantics. Plummer et al~\cite{plummer2015flickr30k} suggest using Canonical Correlation Analysis (CCA) and Wang et al~\cite{wang2016deepcca} suggest Deep CCA to learn similarity among visual and language modalities. Wang et al.~\cite{wang2016structured} employ structured matching and boost performance using partial matching of phrase pairs. Plummer et. al~\cite{plummer2017sppc} further augment CCA model to take advantage of extensive linguistic cues in the phrases. All these approaches are upperbound by the performance of the proposal generation systems.

Recently, Chen et al~\cite{myiccv2017} proposed QRC that overcomes some limitations of region proposal generators by regressing proposals based on query and employs reinforcement learning techniques to punish conflicting predictions.

\textbf{Visual and Semantic context} Context provides broader information that can be leveraged to resolve semantic ambiguities and rank proposals. Hu et al~\cite{hu2016natural} used global image context and bounding box encoding as context to augment visual features. Yu et al.~\cite{yu2016modeling} further encode the size information and jointly predict all query regions to boost performance in a referring task. Plummer et al.~\cite{plummer2017sppc} jointly optimize neighboring phrases by encoding their relations for better grounding performance. Chen et al~\cite{kan2017msrc}~\cite{myiccv2017} employ semantic context to jointly ground multiple phrases to filter conflicting predictions among neighboring phrases. The existing approaches for grounding do not take full advantage of the rich context provided by visual context, semantic context and inter-phrase relationships.

\textbf{Knowledge Transfer} Knowledge transfer involves solving a target task by learning from a different but related source task. Li et al.~\cite{Li2006KnowledgeTI} employ knowledge transfer, Rohrbach et al~\cite{Rohrbach2011KT} use linguistic knowledge bases to automatically transfer information from source to target classes. Deselaers et al~\cite{Des2012Weak} and Rochan et al.~\cite{rochan2015weak} employ knowledge transfer in weakly-supervised object detection and localization respectively using skip vectors~\cite{Mikolov2013W2V} as a knowledge base. In this work, we propose to use knowledge transfer from object detection task to index region proposals for weakly supervised phrase grounding.

\section{Our network}
\begin{figure*}[h]
\vspace*{-0.25cm}
\includegraphics[width=0.95\textwidth,height=2.5in]{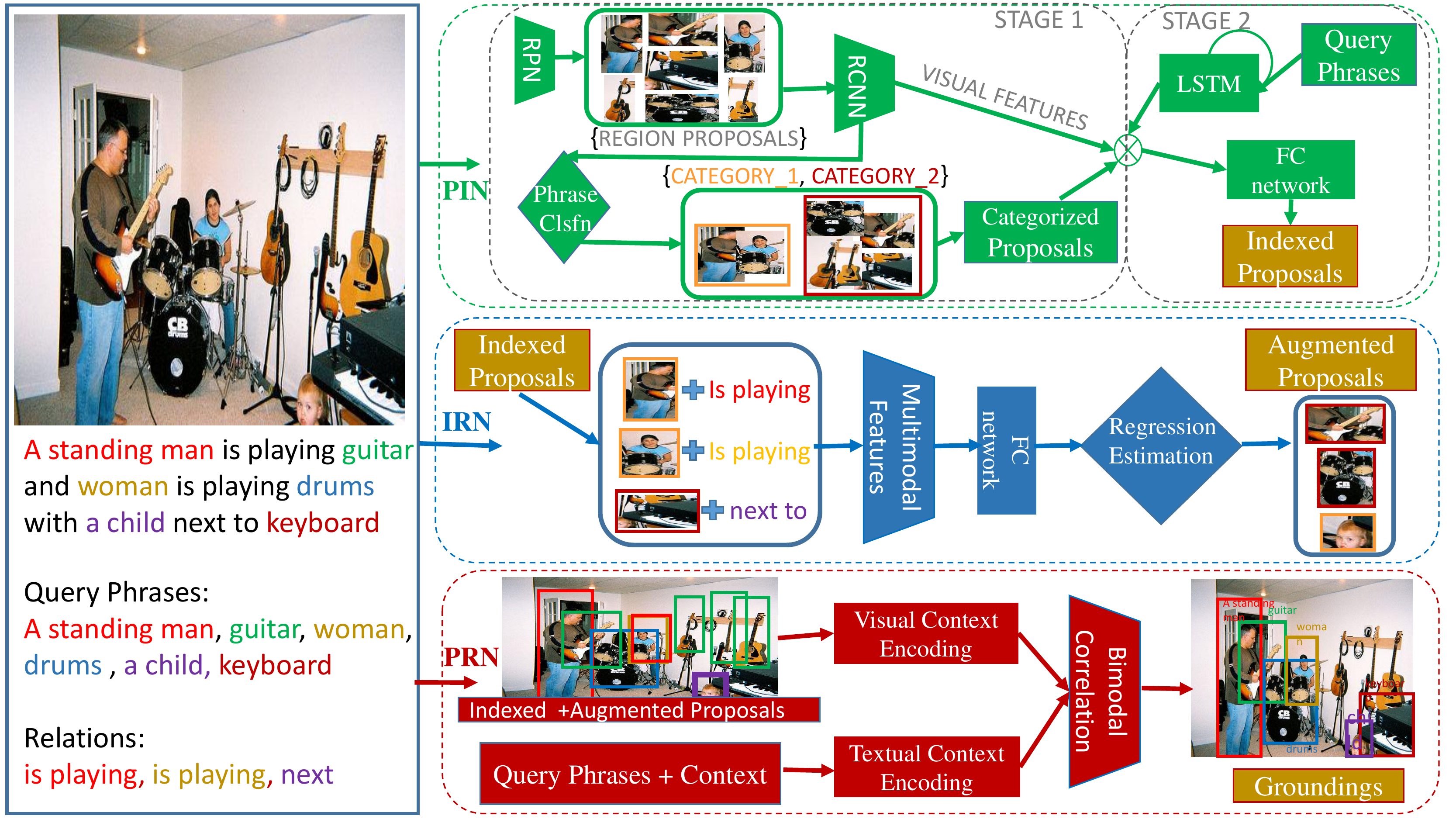}
\centering
\vspace*{-0.25cm}
\caption{Architecture of PIRC Net}\label{fig: arch}
\vspace*{-0.75cm}
\end{figure*}

In this section, we present the architecture of our PIRC network (Figure \ref{fig: arch}). First, we provide an overview of the entire framework followed by detailed descriptions of each of the three subnetworks : Phrase Indexing Network (PIN), Inter-phrase Regression Network (IRN) and Proposal Ranking Network (PRN).
\subsection{Framework Overview}
Given an image $\mathbb{I}$ and query phrases $\mathit{p_i} \in \{\mathbb{C}\}$, the goal of our system is to predict the location of visual entities specified by the queries. PIN takes each query phrase and image as input and in two stages, retrieves a set of ``indexed region proposals". IRN generates the region proposals for each query phrase by predicting its location using its relationship with neighboring query phrases of the same image (Note: This information is only available if multiple query phrases are generated from the same image description). The union of proposals generated by PIN and IRN is referred to as ``candidate proposals". Finally, PRN uses context-incorporated features to rank the candidate proposals and choose the one that is most relevant to the query phrase. 

\subsection{Proposal Indexing Network(PIN)}\label{SPIN}
PIN retrieves a subset of region proposals that are likely correlated to a given query phrase in two stages. Classification is used by PIN to categorize the region proposals in the Stage 1. In Stage 2, PIN ranks the proposals and chooses a subset for which complex analysis is performed in later stages by IRN and PRN subnetworks.

      \textbf{Stage 1} Architecture of stage 1 of PIN is analogous to Faster RCNN~\cite{ren2015faster} and has two subnetworks: A proposal generator (RPN) and a proposal classifier (RCNN). For proposal generation, we finetune a pretrained object detection network to generate region proposals like ~\cite{myiccv2017} instead of object proposals. For classification (RCNN), since each query phrase is distinct from another, we propose to group them into fixed number of higher level phrase categories. To achieve this, we encode phrases as skip-thought vectors~\cite{Kiros2015Skip} and cluster them into a set number of phrase categories $\mathit{C_j}, \mathit{j} \in [1,N]$. Skip-thought vectors~\cite{Kiros2015Skip} employ a data-driven encoder-decoder framework to embed semantically similar sentences into similar vector representations. Given a query phrase $\mathit{p_i}$, it is embedded into a skip-thought vector $\mathit{ep_i}$ and then categorized as follows:
\begin{equation}\label{equ:phr_cat}
\begin{aligned}
	\mathit{p_i} \in \mathit{C_j} :  \mathit{ep_i}\cdot\mu_j >  \mathit{ep_i}\cdot\mu_J, \forall_{J \neq j} J \in [1,N]
\end{aligned}
\end{equation}
 where $\mu_j$ is center of cluster j.
 
	\textbf{Stage 2} In stage 1, classification chooses the region proposals that likely belong to the query phrase category (Eq :~\ref{equ:phr_cat}). In stage 2, these proposals are ranked based on their relevance to the query phrase.
	
	For stage 2, we employ visual attention on each region proposal from stage 1 to rank their similarity to query phrase. For each region proposal, its visual feature is concatenated to the query phrase embedding and these multimodal features are projected through a FC layer network to get a 5 dimensional prediction vector $\mathit{s_p^i}$. The first element of the vector indicates the similarity score between proposal and query embedding and the next four elements indicate the regression parameters of the proposals. Visual features of proposals are obtained from the penultimate layer of the classification network in Stage 1. An LSTM is used to encode a query phrase as an embedding vector. For a region proposal with visual feature $\mathit{vf_j}$ and query feature $\mathit{g_{p_i}}$ generated for a query phrase $\mathit{p_i}$, the loss function is calculated as a combination of ranking $\mathcal{L}_{rnk}$ and regression loss $\mathcal{L}_{reg}$ mentioned below:
\begin{equation}\label{equ:loss_rnk}
\begin{aligned}
\mathcal{L}_{rnk}(\{\mathit{vf_j}\},\mathit{g_{p_i}}) = -log(\mathit{s_p^i[0]})
\end{aligned}
\end{equation}
\begin{equation}\label{equ:loss_reg}
\begin{aligned}
\mathcal{L}_{reg}(\{\mathit{vf_j}\},\mathit{g_{p_i}}) = \frac{1}{4} 
\sum_{k=1}^4\mathit{f}(\|s_p^i[k] - s_{g_i}^j[k]\|)
\end{aligned}
\end{equation}
where $s_{g_i}$ are regression parameters for proposals relative to ground-truth and $\mathit{f}$ is a smooth-L1 loss function. The region proposals $\{\mathit{V_S^i}\}$ with highest similarity to query phrase are chosen as indexed region proposals for further inspection.

While the proposals chosen in PIN have fairly high accuracy, they still do not consider any relative attributes and relationships while ranking the proposals. The next modules, IRN and PRN incorporate inter-phrase relations and context knowledge to improve ranking among these indexed region proposals.

\subsection{Inter-phrase Regeression Network(IRN)}
Inter-phrase Regression Network uses a novel architecture to take advantage of the relationship among two neighboring query phrases (from an image description) to estimate the relative location of a target phrase from a source phrase. Given a phrase tuple of {source phrase, relationship, target phrase}; IRN estimates the regression parameters to predict the location of target phrase given the location of source phrase and vice-versa. To model the visual features for regression, the representation of source phrase must encode not only its visual appearance $\mathit{fv}$ but also its spatial configuration and its interdependence with the target phrase. For example, the interdependence of 'person-in-clothes' is different to that of 'person-in-vehicle' and is dependent on where the 'person' is. To encode the spatial configuration $\mathit{lv}$, we employ a 5D vector that is encoded as $\begin{bmatrix}\frac{x_{min}}{W}, \frac{y_{min}}{H}, \frac{x_{max}}{W}, \frac{y_{max}}{H}, \frac{\Delta x \cdot \Delta y}{W \cdot H} \end{bmatrix}$ where ($\mathit{W,H}$) are width and height of the image respectively. To encode the interdependence of the phrases, we suggest to use the phrase categories(from PIN) of source and target phrases embedded as a one-hot-vector $\mathit{rv}$. The relation between the two phrases is encoded using an LSTM ($\mathit{Er}$); concatenated with visual feature and is projected using a Fully Connected Layer to obtain regression parameters for target phrase location. For a query phrase $\mathit{p_i}$ and its neighboring phrases $\mathit{Np_i}$, the regression $\mathit{Rp_i}$ is estimated as follows:
\begin{equation}\label{equ:irn_reg}
\begin{aligned}
\mathit{Rp_i} = \phi(\mathbf{W}_{rl}(\mathit{fv_k||lv_k||rv_k||Er_i})+\mathbf{b}_{rl}); \mathit{v_k} \in \{V_S^{Np_i}\}
\end{aligned}
\end{equation}
where '$||$' denotes the concatenation operator, $\mathit{p_i}$ is the query phrase whose regression parameters are predicted and $\{V_S^{Np_i}\}$ is the set of region proposals chosen by PIN for neighboring phrases $\mathit{Np_i}$ of query phrase $\mathit{p_i}$. $\mathbf{W}_{rl}$ and $\mathbf{b}_{rl}$ are projection parameters and $\phi$ is the non-linear activation function.

 During training, the regression loss $\mathcal{L}_{rlReg}$ for predicted regression parameters $\mathit{Rp_i}$ is calculated as follows:
\begin{equation}
\begin{aligned}\label{equ:loss_irn}
\mathcal{L}_{rlReg}(\mathit{Rp_i},\mathit{g_{p_i}}) = \sum_{k=1}^4\mathit{f}(\|rp_i[k] - {g_{p_i}}[k]\|)
\end{aligned}
\end{equation} 
given the ground truth regression parameters $\mathit{g_{p_i}}$ calculated from ground truth location $\mathit{g}$ of target phrase $\mathit{p_i}$.

 The proposals $\{\mathit{rV_S^i}\}$ estimated from both the neighboring phrases (subject and object) are added to the proposals generated by PIN 
\begin{equation}\label{equ:aug_prop}
\begin{aligned}
 \{\mathit{V_{S_a}^i}\} \equiv \{\mathit{V_S^i}\} \bigcup \{\mathit{rV_S^i}\}
\end{aligned}
\end{equation}  
  IRN enhances the proposal set and is especially useful for smaller objects which are often missed by proposal generators  (Figure \ref{fig: qual}, query 3). The candidate proposal set$\{\mathit{V_{S_a}^i}\}$ is next passed to the Proposal Ranking Network.

\subsection{Proposal Ranking Network(PRN)}
Proposal Ranking Network (PRN) is designed to incorporate visual and semantic context while ranking the region proposals. PRN employs a bimodal network to generate a confidence score for each region proposal using a discriminative correlation metric. For the visual modality, we employ contrastive visual features to differentiate among the region proposals from the same phrase category. Each proposal is encoded by aggregating its relative appearance with respect to other candidate proposals ( Eq: ~\ref{equ:aug_prop} ). This relative appearance between a given proposal and any other candidate proposal is computed by L2 normalization of the difference in visual features of the two proposals. To compute the feature representation, this relative appearance is aggregated using average pooling for a given proposal as follows:  
\begin{equation}
\begin{aligned}\label{equ:rel_vis}
\mathit{cfv_k} =  \sum_{l \neq k}\frac{fv_k-fv_l}{\left\|fv_k-fv_l\right\|}; \nonumber \mathit{v_k} \in \{\mathit{V_{S_a}^i}\}
\end{aligned}
\end{equation}
For location and size feature representation, we encode relative position and size of a given proposal with respect to all the other candidate proposals. This representation helps in capturing the attributes of a proposal compared with other candidate proposals and is especially helpful in relative references. This feature representation is encoded as a 5 D vector:\\
 $\begin{bmatrix} \frac{[\mathit{x}_{c_k} - \mathit{x}_{c_{Nk}}]}{\mathit{w}_k}, \frac{[\mathit{y}_{c_k} - \mathit{y}_{c_{Nk}}]}{\mathit{h}_k}, \frac{[\mathit{w}_{k} - \mathit{w}_{Nk}]}{\mathit{w}_k}, \frac{[\mathit{h}_{k} - \mathit{h}_{Nk}]}{\mathit{h}_k}, \frac{[\mathit{w}_{Nk} * \mathit{h}_{Nk}]}{[\mathit{w}_k*\mathit{h}_k]} \end{bmatrix}$
  using the relative distance between centers, relative width, relative height and relative size of the given proposal and any other candidate proposal respectively. The final visual representation is the concatenation of all the above representations (Eq : ~\ref{equ:prn_enc_v}).

To encode the text modality (Eq : ~\ref{equ:prn_enc_t}), we concatenate the query embedding, $\mathit{g_{p_i}}$ with embedding, $\mathit{g_{fc}}$ of the entire image description $\mathbb{C}$. 
\begin{align}\label{equ:prn_enc_v}
\mathit{Rv_k} &= \phi(\mathbf{W}_{V}(\mathit{fv_k||cfv_k||lv_k})+\mathbf{b}_{V}) 
\end{align}
\begin{align}\label{equ:prn_enc_t}
\mathit{Rp_i} &= \phi(\mathbf{W}_{T}(\mathit{g_{p_i}}||\mathit{g_{fc}})+\mathbf{b}_{T})
\end{align}
 To compute the cross modal similarity, first the textual representation is projected into the same dimensionality as the visual representation. Then, the discriminative confidence score $\zeta(\mathit{v_k}, \mathit{p_i})$ is computed by accounting for the bias between the two modalities as follows:
\begin{align}\label{equ:prn_proj}
\mathit{Vp_i} = \phi(\mathbf{W}_{P}(\mathit{Rp_i})), \mathit{bp_i} &= \phi(\mathbf{b}_{P}(\mathit{Rp_i}))\\ 
\zeta(\mathit{v_k}, \mathit{p_i}) = \mathit{Vp_i}.\mathit{Rv_k} + \mathbf{b}_{p_i}
\end{align}
To learn the projection weights and bias for both modalities during training, we employ the max-margin ranking loss $\mathcal{L}_{rnk}$ that assigns higher scores to positive proposals. To account for multiple positive proposals in the candidate set, we experiment with both maximum and average pooling to get the representative positive score from the proposals. In our experiments, maximum pooling operator performed better. The ranking loss is formulated below:
\begin{equation}\label{equ:prn_loss}
\begin{aligned}
\mathcal{L}_{rnk} = \sum_{N_k} max[0, \lambda+\zeta(\mathit{v_{N_k}}, \mathit{p_i})-max(\zeta(\mathit{v_{P_k}}, \mathit{p_i}))]
\end{aligned}
\end{equation}
The loss implies that the score of the highest scoring positive proposal, $\mathit{v_{P_k}} \in \{\mathit{V_{S_a}^i}\}$ should be greater than each of the negative proposal $\mathit{v_{N_k}} \in \{\mathit{V_{S_a}^i}\}$ by a margin $\lambda$.
\subsection{Supervised training and Inference}
The proposal generator for PIN is pre-trained using RPN~\cite{ren2015faster} architecture on PASCAL VOC~\cite{pascal-voc-2007} dataset. The fully-connected network of PIN is alternatively optimized with RPN to index the proposals. Stage 2 of PIN is trained independently for 30 epochs with a learning rate of 1e-3. IRN and PRN for are trained for 30 epochs with starting learning rate of 1e-3 that is reduced by a factor of 10 every 10 epochs. During testing, the region proposal with the highest score from PRN is chosen as the prediction for a query phrase.

\section{Weakly Supervised Training}
We present our framework for weakly-supervised training in this section. 

\subsection{Weak Proposal Indexing Network (WPIN)}
For weakly-supervised grounding, to overcome the lack of ground truth information, we employ knowledge transfer learning from object detection systems for indexing relevant proposals for a query phrase. The knowledge transfer is two-fold : data-driven knowledge transfer and appearance based knowledge transfer. We describe both methodologies below.

\textbf{Data-driven Knowledge Transfer}
For Data-driven Knowledge Transfer, our training objective is to learn the representations of phrase categories from a pre-trained object detection network. The pre-trained network's relevant object classes provide a strong initialization for generating proposals of phrase categories (defined similarly as Section \ref{SPIN}). For a region proposal $ \mathit{v_k} \in \{\mathit{V_T}\}$, the network trains to predict probability $\{\mathit{pc_j}\}$ for a phrase category $\mathit{C_j}, \mathit{j} \in [1,N]$ . Final scoring layer of pre-trained network is replaced to predict the probabilities of $\mathit{N}$ phrase categories. Each region proposal $\mathit{v_k}$ is represented as the distribution of the phrase category probabilities $[pc_1^k..pc_J^k]$. For training the network, the representations of $\mathit{K}$ region proposals for an image $\mathbb{I}$ are added and loss function $\mathcal{L}_{dkt}$ is calculated as a multi-label sigmoid classification loss as follows:
\begin{equation}\label{equ:wpin_ddk}
\begin{aligned}
\mathcal{L}_{dkt} = \frac{1}{N}\sum_{j=1}^N \mathit{y_j}*log(\mathit{PC_j})+ (1-\mathit{y_j})*log(1-\mathit{PC_j})
\end{aligned}
\end{equation}
 where $\mathit{PC_j} = \sigma (\frac{1}{K}\sum_{k=1}^K pc_j^k)$. $\sigma$ denotes the sigmoid function. $\mathit{y_i} = 1$ if the image contains phrase category and 0 otherwise. During test time, the region proposals with highest scores $\{pc_j^{max}\}$ for a query phrase $ \mathit{p} \in \mathit{C_j}$, are chosen as the indexed query proposals $\{\mathit{V_I}\}$.
  
\textbf{Appearance-based Knowledge Transfer}
Appearance-based knowledge Transfer is based on the expectation that semantically related object classes have visually similar appearances. While this may not hold true universally and could mislead the system in few cases, it provides a strong generalization among classes that do. Given probability scores $\{\mathit{pc_o}\}$ of a set of source classes $\{\mathit{sc_o}\}$ for a region proposal $\mathit{v_k} \in \{\mathit{V_T}\}$, the goal of the knowledge transfer is to learn the correlation score $\mathit{S_{pk}^j}$ for a query phrase $\mathit{p}$ for that region proposal $\mathit{v_k}$. To measure the correlation among different classes, we employ skip vectors~\cite{Mikolov2013W2V} that embed semantically related words in similar vector representations. For a query phrase $\mathit{p}$, we employ its constituent nouns $\mathit{NP}$ extracted using Stanford POS tagger along with phrase category $\{\mathit{C_j}\}, j \in [1,N]$ for its semantic representation . For a set of $\mathit{K}$ proposals $\{\mathit{v_k}\}$ given by an object detection system with source class probability scores $\{\mathit{pc_o^k}\}$; we measure their correlation $\mathit{S_{pk}^j}$ to target phrase class $\mathit{C_j}$ and phrase's constituent nouns $\mathit{NP}$ as follows:
\begin{equation}\label{equ:wpin_abk}
\begin{aligned}
\mathit{S_{pk}^j} = \frac{\sum_{o=1}^O\mathit{pc_o}*\mathit{Vsc_o}}{||\sum_{o=1}^O\mathit{pc_o}*\mathit{Vsc_o}||} . \frac{V(\mathit{C_j})+V(\mathit{NP})}{||V(\mathit{C_j})+V(\mathit{NP})||} \\ 
\end{aligned}
\end{equation}
An average of appearance-based correlation $\mathit{S_{pk}^j}$ and data-driven probability ${pc_j^k}$ is employed as the final score for correlation of proposal $\mathit{v_k}$ with query phrase $\mathit{p}$.
\subsection{Training and Inference}
Faster RCNN system~\cite{ren2015faster} pretrained on MSCOCO~\cite{lin2014coco} dataset using VGG~\cite{Simonyan14c} architecture is employed for knowledge transfer. For training the weakly supervised grounding system, the encoder-decoder network with attention is used to compute reconstruction loss similar to ~\cite{rohrbach2016grounding}. The learning rate of the network is set to 1e-4.

\section{Experiments and Results}
We evaluate our framework on Flickr30K Entities~\cite{plummer2014flickr30k} and Referit Game~\cite{KazemzadehOrdonezMattenBergEMNLP14} datasets for phrase grounding task.

\subsection{Datasets}
\textbf{Flickr30k Entities}
We use a standard split of 30,783 training and 1000 testing images. Each image has 5 captions, 360K query phrases are extracted from these captions and refer to 276K manually annotated bounding boxes. Each phrase is assigned one of eight pre-defined phrase categories. We treat the connecting words between two phrases in the caption as a 'relation' and use relations occurring $>$ 5 times for training IRN.

\textbf{ReferIt Game}
We use a standard split of 9,977 training and 9974 testing images. A total of 130K query phrases are annotated to refer to 96K distinct objects. Unlike Flick30K, the query phrases are not extracted from a caption and do not come with an associated phrase category. Hence, we skip training IRN for ReferIT.

\subsection{Experimental Setup}
\textbf{Phrase Indexing Network (PIN)} 
A Faster RCNN pre-trained on PASCAL VOC 2007~\cite{pascal-voc-2007} is finetuned on the respective datasets for proposal generation. Flickr30k~\cite{plummer2014flickr30k} and ReferIt Game~\cite{KazemzadehOrdonezMattenBergEMNLP14} employ 10 and 20 cluster centers obtained from clustering training query phrases as target classes respectively. Vectors from the last fc layers are used as visual representation for each proposal. For Stage 2, hidden size and dimension of bi-LSTM are set to 1024. 

\textbf{InterPhrase Regression Network (IRN)}
Since query phrases of ReferIt game are annotated individually, IRN is only applicable to Flickr30k dataset. The visual features from PIN are concatenated with 5D spatial representation (for regression) and 8*8 one-hot embedded vector for source and target phrase categories; for generating representation vector. Both left and right neighboring phrases, if available, are used for regression prediction.

\textbf{Proposal Ranking Network(PRN)}
For visual stream, the visual features from PIN are augmented with contrastive visual features and 5D relative location features; generating an augmented feature vector. For text stream, lstm features are generated for both query phrase and corresponding caption. Each stream has 2 fully connected layers followed by a ReLU non-linearity and Dropout layers with probability \= 0.5. The intermediate and output dimensions of visual and text streams are [8192,4096].

\begin{table}[t]
  \centering
  \begin{tabular}{lc} \toprule
  Approach & Accuracy (\%) \\ \midrule
  \textbf{Compared approaches} & \\
  SCRC~\cite{hu2016natural} & 27.80 \\
  Structured Matching~\cite{wang2016structured} & 42.08 \\
  GroundeR~\cite{rohrbach2016grounding} & 47.81 \\
  MCB~\cite{fukui2016multimodal} & 48.69 \\
  CCA embedding~\cite{plummer2015flickr30k} & 50.89 \\
  SPPC~\cite{plummer2017sppc} & 55.85 \\
  MSRC ~\cite{kan2017msrc} & 57.53 \\
  QRC ~\cite{myiccv2017} & 65.14 \\ \midrule
  \textbf{Our approaches} & \\
  PIN (VGG Net)  & 66.27 \\
  PIN + IRC (VGG Net) & 70.17 \\     
  PIN + PRN (VGG Net) & 70.97 \\ 
  PIRC Net (VGG Net) & \textbf{71.16} \\ \midrule
  PIN (Res Net)  & 69.37 \\
  PIN + IRC (Res Net) & 71.42 \\     
  PIN + PRN (Res Net) & 72.27 \\ 
  PIRC Net (Res Net) & \textbf{72.83} \\
  \bottomrule
  \end{tabular}
  \vspace{1.0mm}
\caption{Relative performance of our apprach on Flickr30k dataset. Different combinations of our modules are evaluated.}\label{tab: flickr30k_res}
  \vspace{-0.65cm}
\end{table}
\begin{table}[t]
  \centering
  \begin{tabular}{lc} \toprule
  Approach & Accuracy (\%) \\ \midrule
  \textbf{Compared approaches} & \\
  Deep Fragments~\cite{karpathy2014deep} & 21.78 \\
  GroundeR~\cite{rohrbach2016grounding} & 28.94 \\
  \midrule
  \textbf{Our approach} & \\ 
  PIRC Net & \textbf{34.28} \\
  \bottomrule
  \end{tabular}
  \vspace{1.0mm}
\caption{Relative performance of our approach for Weakly supervised Grounding on Flickr30k dataset.}\label{tab: flickr30k_wek_res}
  \vspace{-1.0cm}
\end{table}

\textbf{Network Parameters}
All convolutional and fully connected layers are initialized using MSRA and Xavier respectively. All the features are l2 normalized and batch normalization is employed before similarity computation. Training batch size is set to 40 and learning rate is 1e-3 for both flickr30k and referit. VGG architecture~\cite{Simonyan14c} is used for PIN to be comparable to existing approaches. Further, ResNet~\cite{He2016DeepRL} architecture is used to establish the new state-of-the-art with improved visual representation. In the experiments, we use VGG net for comparison with other methods and ResNet for performance analysis. We set $\mathit{L}$ as 10 and 20 for VGG net and ResNet respectively.

\textbf{Evaluation Metrics}
Accuracy is adopted as the evaluation metric and a predicted proposal is considered positive, if it has an overlap of $>$0.5 with the ground-truth location. For evaluating the efficiency of indexing network, Top 3 and Top 5 accuracy are also presented.

\begin{table}[t]
  \centering
  \begin{tabular}{|l|c|c|c|} \hline
  Retrieval Rate & Top 1 & Top 3 & Top 5 \\ \hline
  QRC\footnote[1]~\cite{myiccv2017} & 65.14  & 73.70 & 78.47 \\ \hline
  PIRC Net & \textbf{66.27} & \textbf{79.80} & \textbf{83.46} \\  \hline
  Proposal Limit (no regression) & \multicolumn{1}{c}{}& \multicolumn{1}{c}{} &83.12   \\ \hline 
  \end{tabular}
\caption{Performance evaluation of efficiency of supervised PIN on Flickr30k Entities}\label{tab: retF_comparison}
  \vspace{-1.0cm}
\end{table}
\begin{figure}[h]
\includegraphics[height=1.0in, width=0.95\textwidth]{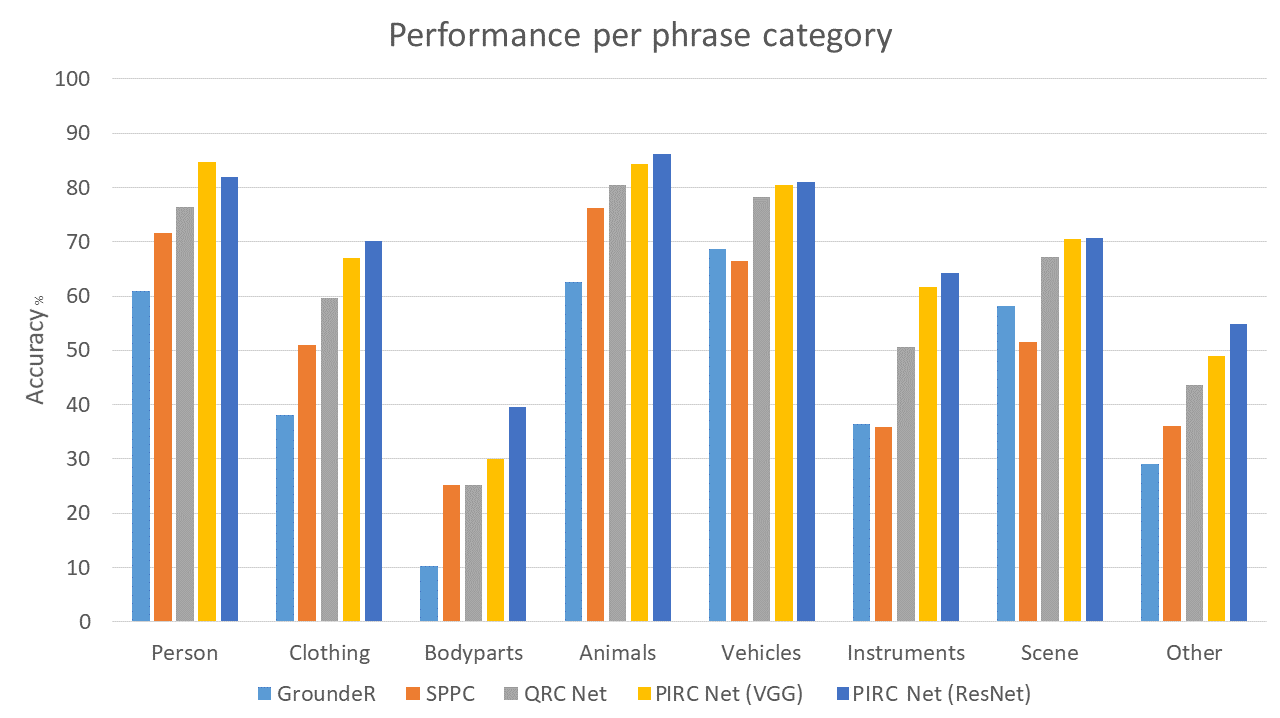}
\vspace{-0.25cm}
\centering
\caption{Grounding performance per phrase category for different methods on Flickr30K Entities}\label{fig: PCperf}
\vspace{-1.25cm}
\end{figure}

\subsection{Results on Flickr30k Entities}
\textbf{Performance}
We compare performance of PIRC Net to other existing approaches for Flickr30k dataset. As shown in table ~\ref{tab: flickr30k_res}, PIN alone achieves 1.13\% improvement while using a fraction of proposals compared to existing approaches. Adding IRN with ranking objective, we achieve 5.05\% increase in accuracy. Adding PRN along with PIN, we achieve 5.83\% improvement. Our overall framework achieves 6.02\% improvement over QRC which is the existing state-of-the-art approach. Further, employing Resnet architecture for PIN gives an additional 1.67\% improvement.

\textbf{Weakly-supervised Grounding Performance}
We compare performance of PIRC Net to two other existing approaches for Weakly-supervised Grounding. As shown in table ~\ref{tab: flickr30k_wek_res}, we achieve 5.34\% improvement over existing state-of-the art.
\begin{table}[t]
  \centering
  \begin{tabular}{lc} \toprule
  Approach & Accuracy (\%) \\ \midrule
  \textbf{Compared approaches} & \\
  SCRC~\cite{hu2016natural} & 17.93 \\
  GroundeR ~\cite{rohrbach2016grounding} & 26.93 \\
  MSRC ~\cite{kan2017msrc} & 32.21 \\
  Context-aware RL~\cite{wu2017refit} & 36.18 \\
  QRC ~\cite{myiccv2017} & 44.07 \\   
  \midrule
  \textbf{Our approaches} & \\
  PIN (VGG Net) & 51.67 \\
  PIRC Net (PIN+PRN) (VGG Net) & \textbf{54.32} \\
  \midrule
  PIN (Res Net)& 56.67 \\
  PIRC Net (PIN+PRN) (Res Net) & \textbf{59.13} \\  
  \bottomrule
  \end{tabular}
  \vspace{1.0mm}
\caption{Relative performance of our approach on Referit Game dataset.}\label{tab: referit_res}
\end{table}
\begin{table}[t]
  \centering
  \begin{tabular}{lc} \toprule
  Approach & Accuracy (\%) \\ \midrule
  \textbf{Compared approaches} & \\
  LRCN~\cite{donahue2015lrcn}(reported in ~\cite{hu2016natural}) & 8.59 \\
  CAFFE-7K~\cite{guad2016caffe}(reported in ~\cite{hu2016natural}) & 10.38 \\
  GroundeR~\cite{rohrbach2016grounding} & 10.69 \\
  \midrule
  \textbf{Our approach} & \\ 
  PIRC Net & \textbf{14.39} \\
  \bottomrule
  \end{tabular}
  \vspace{1.0mm}
\caption{Relative performance of our approach for Weakly supervised Grounding on Referit dataset.}\label{tab: referit_wek_res}
\vspace{-1.15cm}
\end{table}

\textbf{PIN indexing performance}
	The effectiveness of PIN is measured by its ability to index proposals for query phrase. For this, we measure the acurracy of proposals at Top 1, Top 3 and Top 5 ranks. We compare the results to QRC, which is the current state of the art in table ~\ref{tab: retF_comparison}. The upperbound i.e., the maximum accuracy possible with RPN is mentioned in the last row. PIN consistently outperforms QRC~\cite{myiccv2017}\footnote[1]{Results provided by authors} in Top 3 and Top 5 retrieval showcasing its effectiveness in indexing the proposals. Our method employs regression to move proposals to positive locations improving over upperbound.
     
\textbf{Per category performance} For the predefined phrase categories of Flickr30K Entities, PIRC Net consistently performs better across all categories ~\ref{fig: PCperf}. 

\subsection{Results on ReferIt Game}
\textbf{Performance}
Table ~\ref{tab: referit_res} shows the performance of PIRC Net compared to existing approaches. PIN network gives a 7.6\% improvement over state-of-the-art approaches. The high gains could be attributed to richer diversity of objects in ReferIt Game than Flickr30k Entities dataset. Employing PRN in addition to PIN, gives a 10.25\% improvement over QRC. Using ResNet architecture gives an additional 4.81\% improvement; leading to 15.06\% improvement over the state-of-the-art. 

\textbf{Weakly-supervised Grounding Performance}
We compare performance of PIRC Net to two other existing approaches for Weakly-supervised Grounding on Referit. As shown in table ~\ref{tab: referit_wek_res}, we achieve 3.70\% improvement over existing state-of-the art.

\textbf{PIN indexing performance}
Similar to Flickr30k, we do performance analysis to judge the effectiveness of PIN . The results are presented in Table ~\ref{tab: retR_comparison}. PIN consistently performs better for Top 3 and Top 5 retrieval.

 \begin{table}[t]
  \centering
  \begin{tabular}{|l|c|c|c|} \hline
  Retrieval Rate & Top 1 & Top 3 & Top 5 \\ \hline
  QRC\footnote[1]~\cite{myiccv2017} & 44.07  & 54.96 & 59.45 \\ \hline
  PIRC Net & \textbf{51.67} & \textbf{68.49} & \textbf{73.69} \\  \hline
  Proposal Limit  & \multicolumn{1}{c}{}& \multicolumn{1}{c}{} &77.79   \\ \hline 
  \end{tabular}
  \vspace{1.0mm}
\caption{Performance evaluation of efficiency of supervised PIN on Referit Game}\label{tab: retR_comparison}
\vspace{-0.75cm}
\end{table}
\begin{figure*}[h]
\includegraphics[width=0.95\textwidth,height=2.25in]{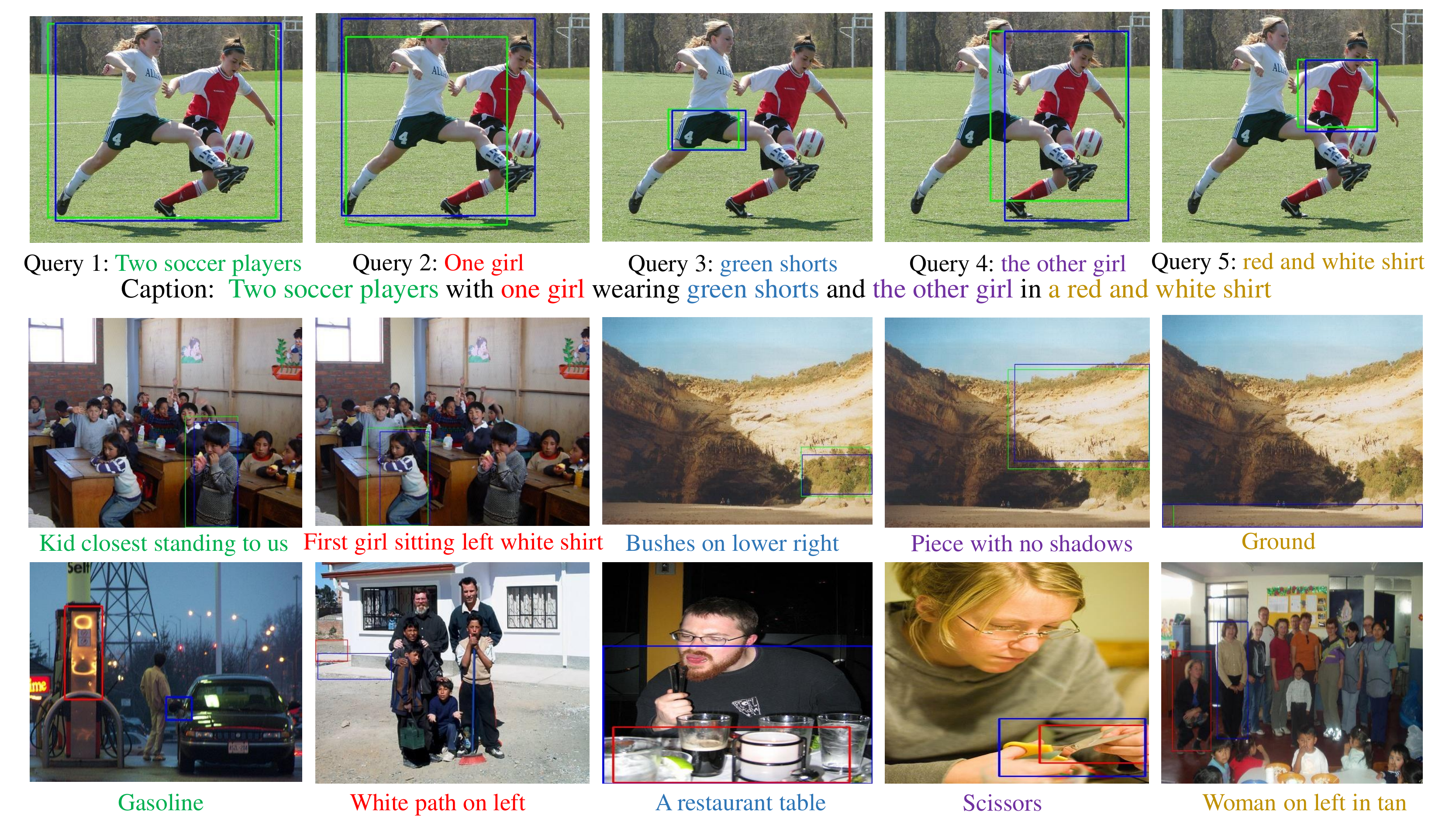}
\centering
\caption{Qualitative results on the test sets of Flickr30K Entities (top row) and ReferIT Game (middle row). Last row shows failure cases from both the datasets. Green: Correct Prediction, Red: Wrong Prediction, Blue: Groundtruth}\label{fig: qual}
\vspace{-0.75cm}
\end{figure*}

\subsection{Qualitative Results}
We present qualitative results for few samples on Flickr30K Entities and ReferItGame datasets(Fig ~\ref{fig: qual}). For Flickr30K (top row), we show an image with its caption and five associated query phrases. We can see the importance of context in localizing the queries. For ReferIt (middle row), we show the query phrase and the associated results. Bottom row shows the failure cases.

\section{Conclusions}
In this paper, we addressed the problem of phrase grounding using PIRC Net, a framework that incorporates semantic and contextual cues to rank visual proposals. By incorporating these cues , our framework outperforms other baselines for phrase grounding. Further, we demonstrate the benefit of knowledge transfer from object detection systems for weakly-supervised grounding.

\section{Acknowledgements}
This paper is based, in part, on research sponsored by
the Air Force Research Laboratory and the Defense Advanced
Research Projects Agency under agreement number
FA8750-16-2-0204. The U.S. Government is authorized
to reproduce and distribute reprints for Governmental purposes
notwithstanding any copyright notation thereon. The
views and conclusions contained herein are those of the authors
and should not be interpreted as necessarily representing
the official policies or endorsements, either expressed
or implied, of the Air Force Research Laboratory and the
Defense Advanced Research Projects Agency or the U.S.
Government.

%
%
%
\bibliographystyle{splncs04}
\bibliography{egbib}
\end{document}